# GENETIC APPROACH FOR ARABIC PART OF SPEECH TAGGING

Bilel Ben Ali[1] and Fethi Jarray[2]

[1]LOGIQ Research Unit, University of Sfax, Tunisia
`bilel_benali@yahoo.fr`
[2] Higher Institute of computer science-Medenine, University of Gabes, Tunisia
`Fethi_jarray@yahoo.fr`

## ABSTRACT

*With the growing number of textual resources available, the ability to understand them becomes critical. An essential first step in understanding these sources is the ability to identify the parts-of-speech in each sentence. Arabic is a morphologically rich language, which presents a challenge for part of speech tagging. In this paper, our goal is to propose, improve, and implement a part-of-speech tagger based on a genetic algorithm. The accuracy obtained with this method is comparable to that of other probabilistic approaches.*

## KEYWORDS

*Part-of-Speech Tagging, Genetic algorithm, Natural Language Processing, Part-of-Speech tagger, Tagset, Training tables & corpus.*

## 1. INTRODUCTION

Arabic is garnering attention in the Natural Language Processing (NLP) community due to its socio-political importance and its linguistic differences from Indo-European languages. These linguistic characteristics, especially dialect differences and complex morphology present interesting challenges for NLP researchers.

Part-Of-Speech Tagging(POST) also known as POS-tagging, word classes, morphological classes, lexical tags or just Tagging is one of the basic tools and components necessary for any robust Natural Language Processing (NLP) infrastructure of a given language. Examples of parts-of-speech are noun, verb, pronoun, preposition, adverb, adjective or other tags. In analyzing the syntactic structure of sentences in a text it is necessary to group words into classes or categories. Part-of-Speech Tagging is a linguistic procedure which attaches word category information to the words in a text.

More formally, given a sentence S, which can be defined as a string of words $w_0, w_1, ..., w_i, ..., wn$ ($n=length\ of\ S$). Part- of-speech tagging is the process of assigning each word $w_i$ of S a corresponding tag from the set of tags.

The result of a part-of-speech tagging maybe in itself is not so interesting, but there are many applications in language technology where this information is useful. For example, POST is considered as one of the basic tools needed in speech recognition, natural language parsing, information retrieval, information extraction, Question Answering, Speech Recognition, Text-to-speech conversion, Machine Translation, Grammar Correction and many more. It is also one of the main tools needed to develop any language corpus.

The genetic algorithm (GA) is a probabilistic search method based on the principles of natural selection. GA is a stochastic search method that has been successfully applied in many real applications of high complexity. A GA is an iterative technique that applies stochastic operators on a pool of individuals (tentative solutions). An evaluation function associates a value to every individual indicating its suitability to the problem. So far, GA has efficiently been used for the solution of combinatorial optimization problems. In this sense, it has been quite successful in many natural language applications for many languages (parsers, part of speech tagger ...). We aim to construct an Arabic Part-Of-Speech tagger with Genetic approach that assigns POS tags to an input text.

The remainder of this paper is organized as follows: The next section is allocated to present a survey of the current approaches in part of speech taggers. In section *3*, we explain the benefits of Genetic Algorithms. In section *4*, we develop a genetic approach on Arabic Part-of-Speech Tagging Problem. In the last section, we present and discuss numerical results.

## 2. TAGGING TECHNIQUES

Several approaches have been used for building POS taggers such as Linguistic, Statistical, Hybrid, Memory and Transformation Based Learning Approach explained below.

### 2.1. Rule-Based Approach

This system is based on finding and correcting errors. During the training period and from a manually tagged training corpus, the system recognizes its own weaknesses and corrects them by constructing a rule base [1] [2]. Two types of rules are used in the tagger Eric Brill [3]:

- Lexical rules: define the label of the word based on its lexical properties.
-  Contextual rules: refine the labeling, that is to say to return to previously assigned labels and correct by examining the local context.

Both types of rules have the form:

- If a word is labeled *A* is in a context *C*, then change it to *B* (contextual rule).
- If a word has lexical property *P*, then assign the label *A* (lexical rule).
The limitations of this approach are that the rule-based taggers are non-automatic, costly and time-consuming.

### 2.2. Probabilistic/Statistical Approach

After *1980*'s Statistical approach [4] [5] [6] came into existence and gained more popularity, this approach requires much less human effort, successful model during the last years Hidden Markov Models and related techniques have focused on building probabilistic models of tag transition sequences in sentence. Results produced by statistical taggers are giving about *95%-97%* of correctly tagged words. However, these statistical part of speech taggers have several potential drawbacks: they are inflexible (use the same strategy for determining the tag of every word), tagging process use only a small amount of information (the bigram method use information of the preceding word) [8].

## 2.3. Hybrid Approach

A combination of both statistical and rule-based methods has also been used to develop hybrid taggers [9] [10]. These seem to produce a higher rate of accuracy. An accuracy of *98%* has been reported by Tapanainen and Voultilainen (*1994*) [11] [12].

## 2.4. Transformation-Based Learning

Transformation-Based Learning (TBL) often also called Brill tagging [13], was introduced by Eric Brill in *1994* and achieved an accuracy of *97.2%* [14] [15] in same corpus outperforming HMM tagger. The learning algorithm starts by building a lexicon that combines the benefits of both rule-based and probabilistic parts-of-speech tagging. Usually the tagger first assigns to every word the most likely part-of-speech. This will introduce several errors. The next step is to correct as many errors as possible by applying transformation rules that the tagger has learned.

## 2.5. Memory-Based Learning

Memory-Based Learning (MBL) [16] [17] [18] is a simple learning method where examples are massively retained in memory. The similarity between memory examples and new examples is used to predict the outcome of a new example. MBL contains two components:

- A learning component which is memory storage.
- A performance component that does similarity-based classification.

# 3. BENIFITS OF GENETIC ALGORITHMS

The search of an optimal solution in GA is heuristic by its nature; heuristic is a rule of thumb that probably leads to a solution. It plays a major role in search strategies because of exponential nature of the most problems. Heuristics help to reduce the number of alternatives from an exponential number to a polynomial number. Possible solutions are suggested and fitness values obtained for the solutions. Then GA, through generations of evolution, provides the possible optimal solutions. In this way, the time complexity of $O(p^n)$ is reduced to $O(P*f(n))$. The function $f(n)$ [19] gives a dependency between the number of possible selections and the generations needed to provide an optimal solution [20]. The integer $n$ is the number of entities in the solution (it is equal to the number of genes in the chromosome). For a sentence of *10* words each with *4* possible tags, in GA, the computation effort will be *10* 20*100 = 20.000*, given a population size of *20* and *100* generations. This compares very favorably to $4^{10}$ times of computations in an exhaustive approach.

GAs search for an optimal solution is very much longer that those of Hidden Markov Model (HMM) and Recurrent Neural Net (RNN). In later cases, the time consuming part is the training step. After training, they normally provide the optimal solution in few times. Typically, a HMM tagger tags more than *100* sentences in one second [21]. Thus it takes least times than the GA tagging process. However, we have to consider the long training time for HMM and RNN. Sometimes, it takes several days and nights to train a RNN. In GA, after obtaining the necessary parameters (training tables, possible tags of each word), it is ready to provide the solution. GA is therefore readily adaptable to the changing inputs whereas, in RNN, it is more difficult to make them adaptable to the change of inputs. HMM can sometimes be as fast, but it is not as flexible as GA in including rules and functions into its computations.

# 4. APPLICATION OF GENETIC ALGORITHM IN TAGGING PROBLEM

Genetic algorithm (GA) is the most popular evolutionary algorithm used successfully in difficult optimization and search problems. It consists of a population of trials, a fitness function to map the genotype trials to real numbered phenotypes, and a set of genetic operators to create new trials.

## 4.1. Preliminary Data Setup

Preparing data for a given tool is an important phase to achieve the goal into consideration. That is why we spent some time on the construction of these data (Hand-Tagged corpora, Tagset and Training Tables).

For the training corpus we are obliged to construct our own corpus because of its non availability for free for Arabic language. We took the EASC corpus [22] that contains many articles talking about the ten following "categories": Art and Music, religion, education, science and technology, environment, sports, finance, tourism, health, politics. To enrich our work and to cover as much as possible frequent words in Arabic we decide to tag some articles from each domain in the watan corpora [23]. Our tagged corpus is constructed as follows: we tag some articles manually, then we execute our tagger taking an article not tagged as an input, then we do the verification manually, we add it to our corpus and so on.

The set of tags contain *22* tags (without punctuation marks) that identify the main tokens in Arabic Language. The choice of these tags was obtained from an adaptation of the tag set English into Arabic during the creation of our corpus. The tag set of this corpus is not too large, what favors the accuracy of the system. Moreover, this tag set has been reduced by grouping some related tags under a unique name tag, what improves statistics.

Training tables are extracted from the corpus, stored into a file, sorted and counted. These tables are extracted as follows: we took to the training corpus and we compute the occurrence of each tag occurred in our training corpus in a gived context. The contexts corresponding to the position at the beginning and the end of the sentences lack tags on the left-hand side and on the right-hand side respectively, this is managed by introducing a special tag, **NULL**, to complete the context. These tables have the following structure:

$$LC_{l_{LC}} \ldots LC_2 \ LC_1 \ \mathbf{T} \ RC_1 \ RC_2 \ldots RC_{l_{RC}}$$

Where $l_{LC}$ the size of left context and $l_{RC}$ the size of right context of the current tag $\boldsymbol{T}$. For example, if $l_{LC} = l_{RC} = 2$, the structure of the training the table could have the following form:

| CN | Prep | Res | Res | CN | 6 |
|----|------|-----|-----|------|-----|
| PN | CN | Prep | Res | Res | 1 |
| Null | PN | CN | Prep | Res | 4 |
| Null | Null | PN | CN | Prep | 135 |
| CN | Prep | V | Null | Null | 105 |

...

We have also constructed a table that records all possible tags of each word occurred in the training corpus. This table stored in a file is useful in the initialization of the first population and in the mutation step, each tag is replaced with a valid tag corresponding to the word that we focus on and this information is extracted from this table.

## 4.2. Individual Representation

Each gene represents a tag. This representation consists of as many genes as words there are in our sentence, and its values will be integer numbers. The chromosome will indicate the tag of each word. This type of coding has an interest to make it possible to create operators of simple crossing and mutation.

## 4.3. Initial Population

For a given sentence of the test corpora, the chromosomes forming the initial population are created by selecting from our training corpus one of the valid tags for each word that appear most frequently in that given context. Words not appearing in the training corpus are assigned the tag which appears more often with that given context in the Hand-Tagged Corpora. Each gene corresponds to each word in the sentence to be tagged. The figure 1 shows some individual examples.

| | المدرسة | إلى | محمد | ذهب |
|---|---|---|---|---|
| Indv1: | Noun | Preposition | Proper Noun | Noun |
| | **0** | **8** | **2** | **0** |
| Indv2: | Noun | Preposition | Proper Noun | Verb |
| | **0** | **8** | **2** | **7** |

Figure 1. Example of individuals

## 4.4. Fitness Function

This function is related to the total probability of the sequence of tags of an individual. The raw data to obtain this probability are extracted from the training table. The fitness of an individual is defined as the sum of the fitness of its genes $(\sum_i f(g_i))$. The fitness of a gene is defined as:

$$f(g) = \log(P(T \mid LC, RC))$$  (1)

Where $(P(T/LC, RC))$ is the probability that the tag of gene $g$ is $T$, given that its context is formed by the sequence of tags $LC$ to the Left Context and the sequence $RC$ to the right. This probability is estimated from the training table as:

$$P(T \mid LC, RC) \approx \frac{occ(LC, T, RC)}{\sum_{T \in T''} occ(LC, T', RC)}$$  (2)

Where $occ(LC, T, RC)$ is the number of occurrences of the list of tags $LC, T, RC$ in the training table, and $T''$ is the set of all tags.

A particular sequence $LC, T, RC$ may not be listed in the training table, because there is insufficient statistics. In these cases we proceed by successively reducing the size of the context, alternatively ignoring the rightmost and then the leftmost tag of the remaining sequence (skipping the corresponding step whenever either $RC$ or $LC$ are empty) until one of these shorter sequences matches at least one of the training table entries or until we are left simply with $T$. In this latter case we take as fitness the logarithm of the frequency with which $T$ appears in the corpus.

For example if we are evaluating the second individual of Figure 1 and we are considering contexts composed of one tag on the left and one tag on the right of the position evaluated, we have the following formula:

$$f(g_1) = \log(\frac{occ(\text{ProperNoun Verb NULL})}{occ(\text{ProperNoun T' NULL})}) \tag{3}$$

$$f(g_2) = \log(\frac{occ(\text{Preposition ProperNoun Verb})}{occ(\text{Preposition T' Verb})}) \tag{4}$$

For the others genes with the same manner, where occ represents the number of occurrences of the context and T' the set of all possible tags.

## 5. GENETIC OPERATORS

### 5.1. Operator of Selection

This operator is responsible for defining what are the individuals in the population P that will be duplicated in the new population P' and will serve as parents (applying the crossover operator). There are several methods of selection. We use roulette-wheel selection of Goldberg (*1989*) [24] [20]. With this method each individual has a chance of being selected proportional to its performance, so more individuals are adapted to the problem, the more likely they are to be selected.

### 5.2. Operator of Crossing

The purpose of the crossing is to enrich diversity of the population by handling the structure of the chromosomes. Classically, the crossings are considered with two parents and generate two children. Initially, to apply uniform crossover, two individuals are selected with a probability proportional to their fitness, crossover is applied with a probability given by the crossover rate. Then a binary array called mask crossover is filled with random values. This mask is intended to know for each locus, of which parent the first child should inherit the gene therein; if faced with the mask has a locus a *0*, the child inherit the gene therein of parent *1*, if it has a *1* he will inherit a parent's *2*. The creation of child *2* is symmetrically, thus dividing both individuals in two parts which produce two children *C1* and *C2* (see Figure 2).

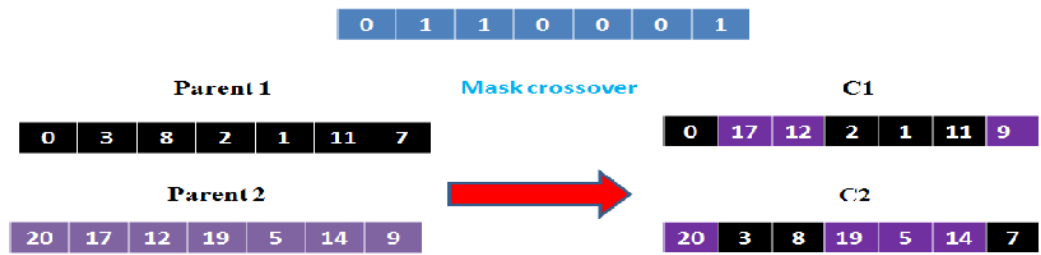

Figure 2.  Uniform crossover

## 5.3. Operator of mutation

The operator of mutation brings to the genetic algorithms the property of ergodicity of course of space. This property indicates that the genetic algorithm will be likely to reach all the points of the space of state, without to traverse them all in the process of resolution. Mutation is then applied to every gene of the individuals resulting from the crossover operation with a probability given by the mutation rate. The number of individuals that remain unchanged in one generation depends on the crossover and mutation rates.

Finally the tag of the mutation point is replaced by another of the valid tags of the corresponding word.

## 6. Tagging Experiments

In this section, we will explain our experimental results on part-of-speech tagging. Direct comparison of performance between researchers is difficult, compounded by variance in corpora, tagset and grading criteria. To calculate Tagging Accuracy Rate (TAR) we use the following formula:

$$TAR = \frac{\text{Number of segments tagged correctly}}{\text{Number of all segments of reference file}} \tag{5}$$

We will measure the impact of many parameters related to our implemented tagger.

## 6.1. Influence of the size of training corpus:

Figure 3 shows Tagger accuracy rate as a function of training corpus size. This curve was generated by training on successive portions of our training corpus. The curve indicates that performance benefit can be obtained by increasing training set size. The Figure 3 illustrates the performance when using different sizes of corpus and it concludes that with the increase in the size of the corpus, the performance of the tagger also increases.

If we train the tagger on large amount of data we get accurate tagging results, where the reason behind this is that when we increase the corpus size we get more combination of tags, which leads to a variety of contexts in training tables that cover various aspects of multiple tags. Thus it concludes that results are dependent on fraction of training data used to train the Tagger. Therefore considering the sizes of corpus used for the experiments, our tagger achieved remarkable accuracy with a limited corpus.

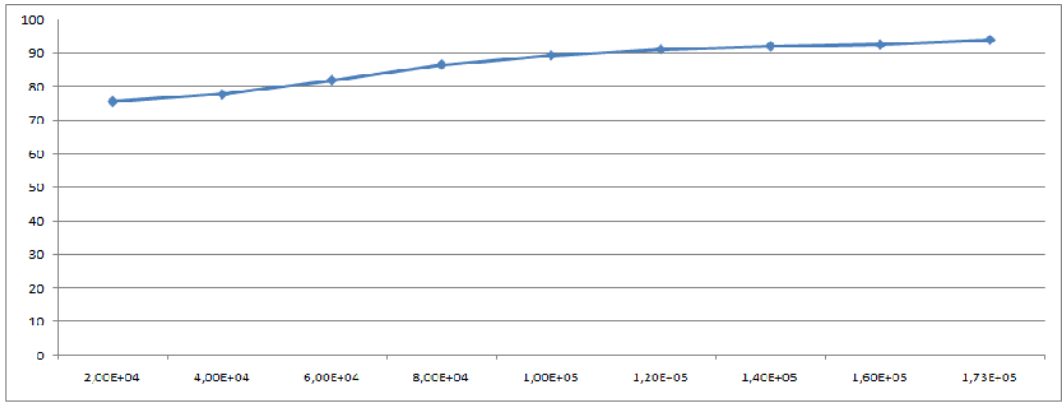

Figure 3. TAR for different sizes of training corpus, a population size of *60* individuals, *30* iterations, a crossover rate of *50%,* and a mutation rate of *4%.* X-axis represents the size of the training corpus. Y-axis represents the tagging accuracy rate.

## 6.2. Influence of Context Size:

The context information has a big influence on the performance of our tagger. The way in which this information is used can increase or decrease the TAR. This information can be varied and have different sizes. Usually two or three preceding words and one or two succeeding words is sufficient. For us we use context size *1-1*, *2-2* and *3-2*.

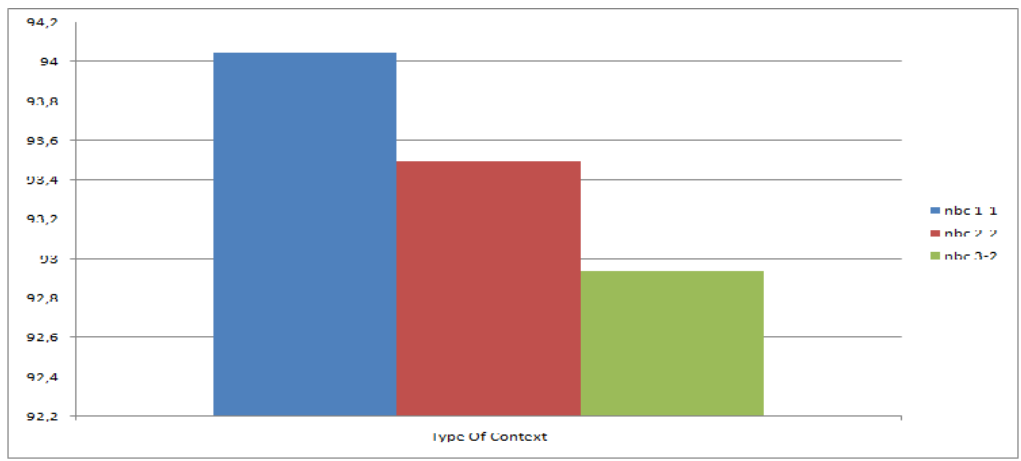

Figure 4. TAR for different sizes of context, a population size of *60* individuals, a crossover rate of *50%,* and a mutation rate of *4%*

The figure 4 shows that the small context size 1- 1 gives best results, and we can explain this by that there are many entries of training tables with larger contexts are not significants. It may be noted also that large context requires more time due to the growing of the size of training tables.

## 6.3. Influence of population size:

The TAR increase with the increasing of population size but we found that population with small size are enough to obtain high accuracy. This can be explained by: the sentences are tagged one by one and so in general a small population is enough to represent the variety of possible taggings. So we are not obliged to start our tagger with a very big population size that does minimize the time of convergence. If population size is too large the GA will take too long time to converge.

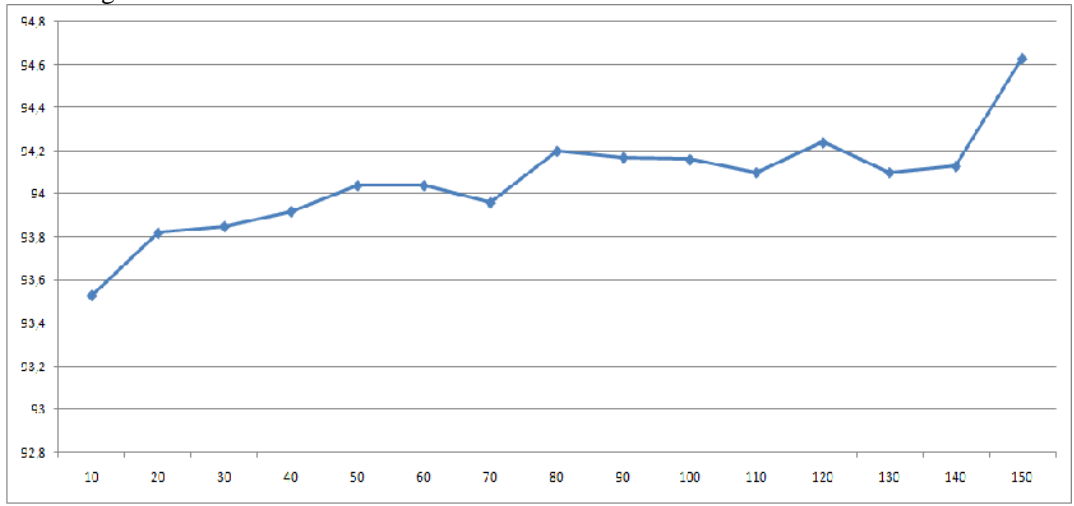

Figure 5.  TAR for different sizes of population, 30 iterations, a crossover rate of 50%, and a mutation rate of 4%

## 6.2. Influence of mutation rate

To prevent to evolve into a local maximum the mutation rate of a Genetic Algorithm is very influential. Many mutation settings have been tried, but only four were eventually reported. The other values do not show a completely different way in the evolutionary process.

During the testing of these variables the same training sets were used. As can be seen in Figure 6, the *5%* mutation rate is most constant over the generations. The results clearly show that a rate of about *0.05* is the best probability to let an individual mutate. This will create enough diversity and prevents the algorithm to grow to a local maximum. This is most desirable because sudden changes in the algorithm could offset it in a wrong way.

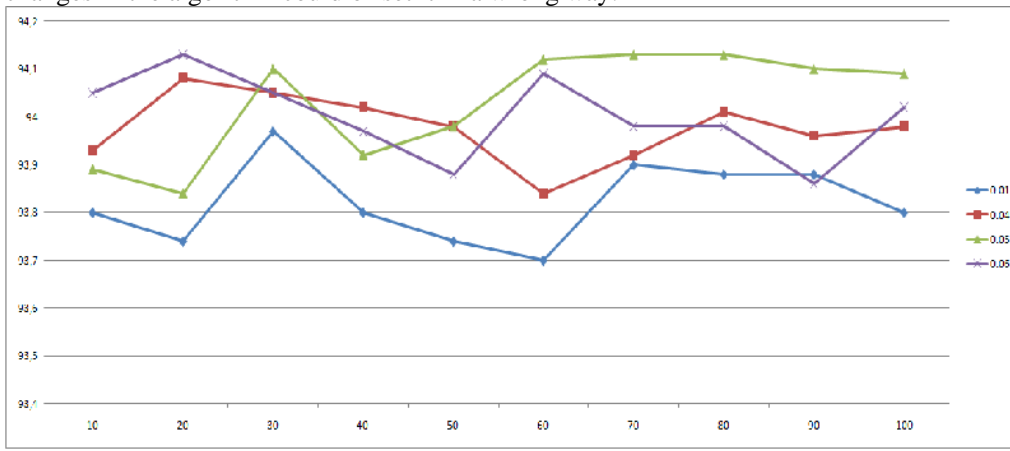

Figure 6.  TAR for different mutation rates, *30* iterations and a crossover rate of *50%*

### 6.3. Influence of crossover parameter

We found on the figure 7 the effectiveness of uniform crossover face one point crossover. In 1-point crossover, the crossover site is randomly selected and in uniform crossover the crossover mask is generated randomly. Uniform crossover is more disruptive in general, it also has a capacity of combining more schemata than 1-point crossover , and this explains results shown in the figure 7.

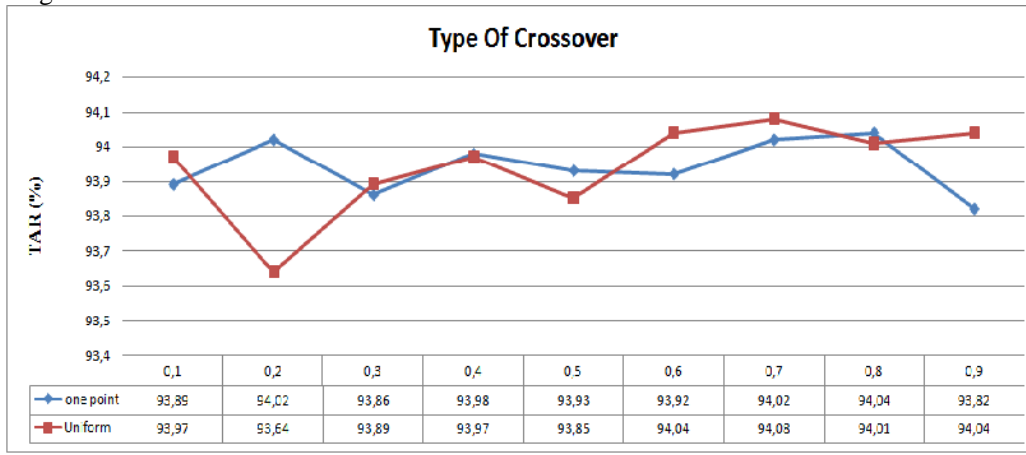

| | 0,1 | 0,2 | 0,3 | 0,4 | 0,5 | 0,6 | 0,7 | 0,8 | 0,9 |
|---|---|---|---|---|---|---|---|---|---|
| one point | 93,89 | 94,02 | 93,86 | 93,98 | 93,93 | 93,92 | 94,02 | 94,04 | 93,82 |
| Uniform | 93,97 | 93,64 | 93,89 | 93,97 | 93,85 | 94,04 | 94,08 | 94,01 | 94,04 |

Figure 7.  TAR Comparison of uniform crossover and crossover on one point, a population size of *45* individuals, *30* iterations, and a mutation rate of *5%*

Varying the rate of recombination independent of the mutation rate has large effect, indicating that recombination of individuals was having a large effect on the performance of the algorithm. The test set accuracies are shown in the figure 8. Results indicate that when crossover is used with *80%* the greatest increase in accuracy is achieved.

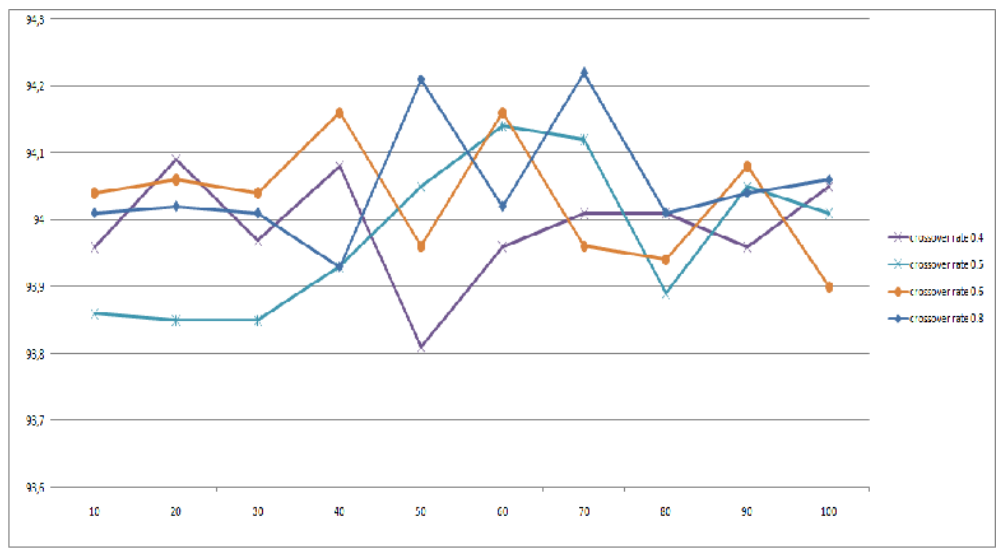

Figure 8.  Influence of crossover rate, a population size of *45* individuals, *30* iterations, and a mutation rate of *5%*

## 7. CONCLUSIONS

We have presented a Genetic approach for Arabic POS tagging that does not suppose segmentation, which would be unrealistic for Arabic language. The approach is competitive although it uses a reduced POS tagset wich allows to obtain interesting results. However, we cannot make direct comparisons with another approach due to the unavailability of standardization of the division of data set into training and test data. The results of our experiments suggest that the Genetic algorithm is a robust enough approach for tagging texts of natural language, obtaining accuracies comparable to other statistical systems. The tests indicate that the length of the contexts surrounding each word extracted for the training and the size of training corpora are determining factors for the results, also our tagger is based on the genetic algorithm which is susceptible to local maximum. Crossover, mutation, together with fitness proportionate selection, serves as a global strategy which can redirect the genetic algorithm to other areas of the search space and grow to the global maximum.

Other possible future works can be reached with genetic approach such as parsing and many other different Natural Language Processing (NLP) tasks.

**Authors**

**Bilel Ben Ali**: Phd student,
 LOGIQ Resarch Unit,
University of Sfax, Tunisia.

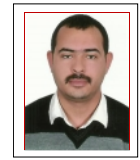

**Fethi Jarray**: Higher Institute of  computer science-Medenine,
University of Gabes, Tunisia.